\documentclass{article}
\usepackage{spconf,amsmath,amssymb,graphicx,booktabs,microtype}
\newcommand{\obj}[1]{\textsc{#1}}  % names of the objectives

\title{TEACHING PPG HOW, NOT WHO: \\
       FIXED-EFFECTS DISTILLATION FROM ECG}

\name{\begin{tabular}{c}Zhongli Wu$^{1}$, Zhuangzhi Gao$^{2}$, Yuankai Wang$^{1}$, Gregory Y.H. Lip$^{1}$, Bilal H. Kirmani$^{4}$, Yalin Zheng$^{3}$\sthanks{Corresponding author: Yalin.Zheng@liverpool.ac.uk}\end{tabular}}

\address{\small $^{1}$Liverpool Centre for Cardiovascular Science, University of Liverpool, UK \\
         \small $^{2}$Shanghai Artificial Intelligence Laboratory, Shanghai, China \\
         \small $^{3}$Department of Eye and Vision Sciences, University of Liverpool, UK \\
         \small $^{4}$Liverpool Heart and Chest Hospital, Liverpool, UK}

\begin{document}
\ninept
\maketitle

\begin{abstract}
ECG is widely used to teach PPG-only models, yet what it teaches is
unexamined. Wearables are valued for tracking how a person's cardiovascular state
changes, but ECG-to-PPG distillation mostly learns who the person is. A
per-recording mean, the trait, holds \mbox{$40$--$59\%$} of a frozen ECG teacher's
target, and pooled students memorise it without carrying it to new recordings.
The raw alignment cosine misses this, since a constant predictor
scores $0.793$. Across 34 runs, the more identity a student memorises, the less
state it learns. Fixed-effects distillation subtracts each recording's mean from
prediction and target, so the trait cancels exactly, while a
pooled anchor keeps it. State agreement more than doubles, within-person labels
improve while age and sex do not, and the gain holds on two
backbones and two further databases. Conditioning on the recording turns
distillation toward the within-person changes that wearables monitor.
\end{abstract}


\section{Introduction}
\label{sec:intro}

Wearables are worn to follow one person over time. What makes a PPG trace useful
is often less how it differs from other people's than how it has changed since an
hour ago, such as a blood-pressure trend or a haemodynamic response during surgery. The
per-person offset is the hard part, which is why cuff-less blood-pressure
estimation is calibrated per user~\cite{Mukkamala_2015}. ECG carries much of the
relevant information, and a growing body of work uses synchronous ECG as a
privileged training signal for PPG-only models: a PPG encoder is aligned to an
ECG model, trained to reconstruct ECG, or coupled to ECG-linked
codes~\cite{Nie_2026,li2026bioxbridge,zhou2026physiologyaware,wu2026p2evqecglinkedrepresentationaugmentation}. Paired ECG does improve PPG-only
representations~\cite{wu2026p2evqecglinkedrepresentationaugmentation}, but which part of the ECG information the alignment
transfers has not been examined.

Much of it, we find, is identity: the part a deployment calibrates away, and the
part that does not carry to new people. Every recording, one monitoring session
of one person, contributes many windows, so a frozen ECG teacher's embedding
splits into a per-recording mean, the \emph{trait}, and a within-recording
deviation, the \emph{state}; the trait holds $40$--$59\%$ of the target energy.
A student that regresses the pooled target fits the trait on training recordings
by reading identity-revealing PPG morphology, and does not carry it to unseen
ones; an InfoNCE whose negatives come from other recordings, as is conventional
for biosignals~\cite{pmlr-v139-kiyasseh21a}, only has to tell people apart. The usual metric does
not reveal this: a constant predictor already reaches a raw cosine of $0.793$
with the target, against $0.823$ for the best linear model.

The problem has a classical form. Each recording's mean is an incidental
parameter in the sense of panel data~\cite{Andersen_1970}, and the
fixed-effects remedy, conditioning on the unit~\cite{Chamberlain_1980}, carries
over directly:
we subtract the recording's mean from prediction and target before comparing
them, so the trait cancels exactly, and keep a pooled anchor that retains it. The change needs
no new architecture and no extra data.

Our contributions are: (i) a diagnosis: pooled distillation learns who rather
than how, the raw alignment cosine cannot tell the two apart, and the identity a
student memorises trades off against the state it learns; (ii) fixed-effects
distillation, a state loss conditioned on the recording under which the trait
cancels exactly, with a pooled anchor that keeps it; and (iii) a within-recording
protocol and evidence: state agreement more than doubles through the
objective, not the sampler, each loss term does its own job, within-person labels
improve while fixed ones do not, and the gain holds on two other databases.

\begin{figure*}[t]
  \centering
  \includegraphics[width=\textwidth]{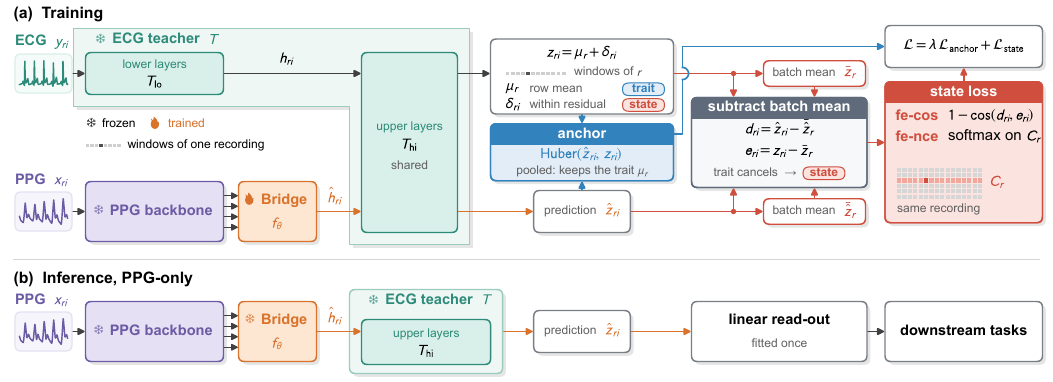}
  \caption{(a) Training. The frozen ECG teacher $T$ is split after its fourth stage: its lower layers
  $T_{\mathrm{lo}}$ map ECG to $400$-channel tokens $h_{ri}$, and the bridge $f_\theta$
  predicts these tokens as $\hat h_{ri}$ from multi-scale features of the frozen PPG backbone. The upper layers
  $T_{\mathrm{hi}}$ map $h_{ri}$ and $\hat h_{ri}$ to $z_{ri}$ and $\hat z_{ri}$. The
  anchor compares raw embeddings, and with equal weight the logits of the teacher's frozen classification head, so
  it keeps the trait $\mu_r$; the state loss compares deviations from the mean over the windows of recording $r$ in
  the batch, where $\mu_r$ cancels, with candidates $C_r$ from the same recording. (b) Inference needs PPG only
  (about $39$\,M parameters with PaPaGei-S, $28.6$\,M of them in $T_{\mathrm{hi}}$); a linear read-out fitted once maps
  $\hat z_{ri}$ to downstream tasks.}
  \label{fig:method}
\end{figure*}

\section{What distillation transfers}
\label{sec:method}

\subsection{Setup and a decomposition of the teacher target}
\label{sec:setup}

Let $r$ index a recording and $i$ a 10\,s window within it. Each window carries a
PPG segment $x_{ri}$ and a synchronous ECG segment $y_{ri}$. A frozen single-lead
ECG foundation model $T=T_{\mathrm{hi}}\circ T_{\mathrm{lo}}$ supplies the target
$z_{ri}=T_{\mathrm{hi}}(h_{ri})\in\mathbb{R}^{D}$, $h_{ri}=T_{\mathrm{lo}}(y_{ri})$.
A bridge $f_\theta$, the only trained module, sees PPG only: it predicts
$\hat h_{ri}=f_\theta(x_{ri})$ from frozen PPG-backbone features. The student is
the whole PPG-only path (backbone, bridge, and the teacher's frozen upper layers
reused as a decoder), and its output is the prediction
$\hat z_{ri}=T_{\mathrm{hi}}(\hat h_{ri})$; ECG is used in the loss and is never
required at inference.

\begin{figure}[t]
  \centering
  \includegraphics[width=\columnwidth]{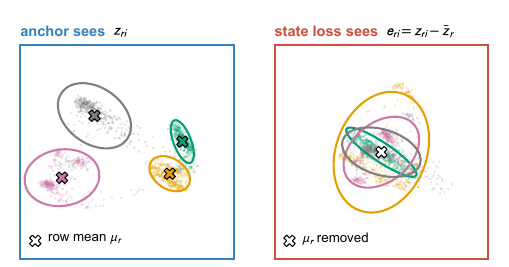}
  \caption{Teacher targets of four training recordings (360 windows each); no model
  is involved. Each panel uses its own principal axes, both at one scale, and each
  ellipse covers two standard deviations of one recording. Left: the anchor compares
  raw targets, where the row means $\mu_r$ ($\times$) set the recordings apart; the
  trait holds $59\%$ of the centred energy, here as over all training recordings.
  Right: the state loss compares $e_{ri}$, where every row mean moves to the origin
  and only the shape of each recording's own variation remains. The pink recording
  visits two states, and its $\mu_r$ lies between them.}
  \label{fig:pca}
\end{figure}

Because every recording contributes many windows, the target splits into a
recording-level and a within-recording part,
\begin{equation}
z_{ri}=\mu_r+\delta_{ri},\qquad
\mu_r=\mathbb{E}_i\!\left[z_{ri}\mid r\right],\qquad
\mathbb{E}_i\!\left[\delta_{ri}\mid r\right]=0 .
\label{eq:decomp}
\end{equation}
We call $\mu_r$ the \emph{trait} component (who is recorded) and $\delta_{ri}$
the \emph{state} component (how they are doing). The trait holds $59\%$ of the centred target energy in the teacher's
raw coordinates and $40\%$ in the standardised coordinates used for training
(282 training recordings; Fig.~\ref{fig:pca}). The state is what changes while the sensor stays on
the same person, and it is what continuous monitoring needs.

\subsection{Why pooled objectives learn identity}
\label{sec:why}

\textbf{The raw metric is dominated by the recording level.} The shared mean and
the per-recording offsets dominate the raw target: the constant predictor (the
training mean of $z$) reaches a raw cosine of $0.793$ against the true target,
and the best linear model only $0.823$, so the usable
range is about $0.03$.

\textbf{Pooled students memorise the trait.} $\mu_r$ is $D$-dimensional and
appears once per recording, so the number of unknowns grows with the data, as
for an incidental parameter~\cite{Andersen_1970}. Minimising
$\lVert\hat z-z\rVert^2$ forces the student to fit it, and on training
recordings it succeeds by reading identity-revealing PPG morphology: with the
batch composition a cross-recording objective would use, trait agreement reaches
$0.90$ on training windows but only $0.33$ on held-out recordings.

\textbf{The two levels need different maps.} A pooled least-squares fit is a
matrix-weighted average of the between- and within-recording
maps~\cite{Maddala_1971}, so it is a compromise whenever they differ. Here they
do: their coefficient cosine is $0.15$ (PaPaGei-S) and $0.25$ (AnyPPG). On
PaPaGei-S a map fitted between recordings predicts trait nearly as well as the pooled
fit ($0.31$ vs $0.34$ on held-out recordings) but reaches only $0.11$ on state,
against $0.26$ for a map fitted within recordings.

\begin{figure*}[t]
  \centering
  \includegraphics[width=\textwidth]{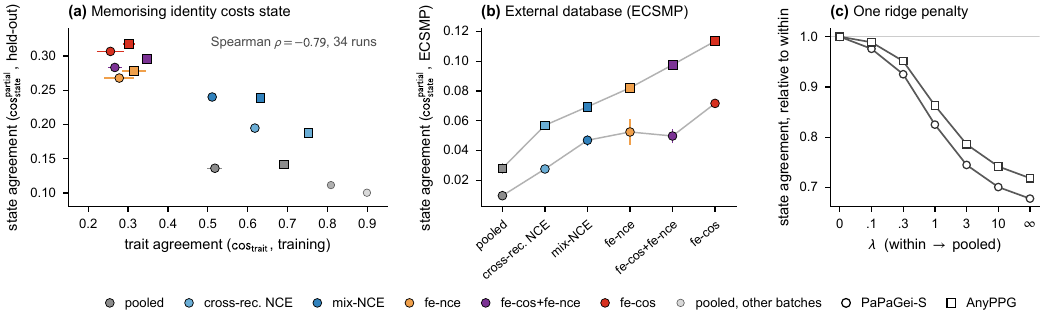}
  \caption{Colour marks the objective and marker shape the backbone in every panel.
  (a) Every run: trait agreement on training windows, which gauges memorised
  identity, against state agreement on held-out recordings; seed mean $\pm$ sd over three
  seeds where available (Sec.~\ref{sec:setup-exp}), and in grey the pooled loss on batches of
  $96\times1$ and $24\times4$ windows. (b) The same objectives on ECSMP, an
  external database with chest ECG and wrist PPG (seed mean $\pm$ sd). (c) The $\lambda$
  path of the linear analogue on ECSMP with one ridge penalty held fixed along
  the path, relative to the within estimator; both backbones decay monotonically
  towards the pooled solution.}
  \label{fig:main}
\end{figure*}

\textbf{Cross-recording negatives reward identity.} An InfoNCE whose negatives
come from other recordings can be solved by identity alone; at the same batch
composition it memorises the trait most strongly of all objectives (Fig.~\ref{fig:main}a).

\section{Conditioning the loss on the recording}
\label{sec:fe}

We condition the state term on the recording, which removes $\mu_r$, and keep a
pooled anchor that carries it (Fig.~\ref{fig:method}):
\begin{equation}
\mathcal{L}=\lambda\,\underbrace{\mathrm{Huber}(\hat z,z)}_{\text{anchor}}
\;+\;\mathcal{L}_{\text{state}} .
\label{eq:loss}
\end{equation}
The anchor is taken on the standardised embedding and, with equal weight, on the
logits of the teacher's frozen classification head. Writing
$d_{ri}=\hat z_{ri}-\bar{\hat z}_r$ and $e_{ri}=z_{ri}-\bar z_r$ for the
deviations from the mean over the windows of recording $r$ present in the batch,
the state term is
\begin{align}
\text{\obj{fe-cos}}: &\quad 1-\cos(d_{ri},e_{ri}), \label{eq:fecos}\\
\text{\obj{fe-nce}}: &\quad -\log\frac{\exp(s_{ii}/\tau)}{\sum_{j\in C_r}\exp(s_{ij}/\tau)},
\label{eq:fence}
\end{align}
where $C_r$ holds the windows of the anchor's own recording, $i\in C_r$, and
$s_{ij}=\cos(d_{ri},e_{rj})$. \obj{fe-cos} is our objective; the conditional
logit \obj{fe-nce} is reported as a variant, in the unweighted sum
\obj{fe-cos}$+$\obj{fe-nce}.

\textbf{Why the trait cancels.} Adding a constant $c_r$ to every target of
recording $r$ leaves $e_{ri}$ unchanged, so $\mu_r$ never reaches either state
term, and it would drop out of Eq.~\ref{eq:fence} even without
centring: with inner-product logits $\langle\hat z_{ri},z_{rj}\rangle/\tau$ over
candidates from one recording, the softmax is unchanged when all logits shift by $\langle\hat z_{ri},c_r\rangle/\tau$. This is how a
conditional likelihood eliminates incidental
parameters~\cite{Andersen_1970,Chamberlain_1980}. Restricting the comparison to
one subject follows the subject-aware contrastive loss of~\cite{cheng2020subjectawarecontrastivelearningbiosignals} and
conditional contrastive learning~\cite{ma2022conditionalcontrastivelearningimproving}; we apply it to a frozen
cross-modal teacher and pair it with an anchor that keeps the removed component.

\textbf{Role of the anchor.} In the linear case, let $X,Y$ stack the inputs and
targets of all windows, $X_w,Y_w$ their within-recording deviations, and $W$ a
coefficient matrix. Minimising
$(1+\lambda)^{-1}(\lVert Y_w-X_wW\rVert_F^2+\lambda\lVert Y-XW\rVert_F^2)+\alpha\lVert W\rVert_F^2$
moves from the within ridge at $\lambda=0$ to the pooled ridge as
$\lambda\to\infty$, with the penalty $\alpha$ keeping the same weight relative to
the data term throughout. In the network the two terms split the work
(Table~\ref{tab:anchor}): without the anchor the trait vanishes but most of
the state gain remains, and the anchor restores the trait and steadies the state.
\begin{table}[t]
\centering
\caption{Held-out agreement on PaPaGei-S without and with the anchor (three seeds).}
\label{tab:anchor}
\vspace{2pt}
\footnotesize
\setlength{\tabcolsep}{4pt}
\begin{tabular}{@{}llccc@{}}
\toprule
 & & state agreement $\uparrow$ & trait agreement & raw cosine \\
\midrule
pooled &  & 0.136{\tiny$\pm$0.006} & 0.316{\tiny$\pm$0.003} & 0.798{\tiny$\pm$0.002} \\
\addlinespace[2pt]
\obj{fe-nce} & w/o anchor & 0.210{\tiny$\pm$0.021} & 0.003{\tiny$\pm$0.002} & 0.109{\tiny$\pm$0.073} \\
 & with anchor & 0.268{\tiny$\pm$0.000} & 0.220{\tiny$\pm$0.031} & 0.797{\tiny$\pm$0.005} \\
\addlinespace[2pt]
\obj{fe-cos} & w/o anchor & 0.249{\tiny$\pm$0.012} & 0.013{\tiny$\pm$0.007} & 0.174{\tiny$\pm$0.040} \\
 & with anchor & 0.307{\tiny$\pm$0.001} & 0.219{\tiny$\pm$0.026} & 0.795{\tiny$\pm$0.007} \\
\bottomrule
\end{tabular}
\end{table}

\section{Experiments}
\label{sec:exp}

\subsection{Data, backbones, and protocol}
\label{sec:setup-exp}

We train on 282 VitalDB recordings from PulseDB~\cite{Wang_2023} (10\,s windows at
125\,Hz, synchronous finger PPG and ECG, intraoperative) and select on 174
held-out recordings; 75 further recordings are reported but never selected on.
The teacher is the frozen single-lead ECGFounder~\cite{Li_2025}. We use two
frozen PPG backbones: PaPaGei-S~\cite{pillai2025papagei}, which has never seen ECG, and
AnyPPG~\cite{Nie_2026}, which was itself aligned to the same teacher, so it shows
whether the effect persists after such alignment. The bridge is a
4.7--4.8\,M-parameter cross-attention module that predicts the teacher's stage-4
tokens from multi-scale backbone features (Fig.~\ref{fig:method}a); the teacher
embedding has $D=1024$ dimensions. All runs use 12\,k steps, batches of 6
recordings $\times$ 16 windows, and three seeds for every objective except cross-recording negatives, with
$\tau=0.1$, a smooth-$\ell_1$ Huber of unit cutoff, and $\lambda=1/9$ for
\obj{fe-cos} and $\lambda=1$ otherwise.

\textbf{State agreement.} Since the raw cosine is saturated
(Sec.~\ref{sec:why}), we remove each recording's own mean from prediction and
target and, as the primary metric, also partial heart rate, its square, and
eight signal-quality descriptors out within each recording
($\cos^{\mathrm{partial}}_{\mathrm{state}}$). Trait agreement
$\cos_{\mathrm{trait}}$, the cosine between predicted and true recording means,
is reported separately; on training windows it gauges memorisation.
Because this metric is close to what \obj{fe-cos} optimises,
Sec.~\ref{sec:what} turns to read-outs that no objective was trained on.

\begin{table*}[t]
\centering
\caption{Public labels and a second database. Linear read-outs of the predicted embedding, fitted on training recordings
and scored on 75 held-out recordings never used for selection and on 60 MIMIC-III recordings the bridge never saw. Errors of
labels that vary within a recording are taken after removing each person's mean; `none' is the no-information reference.
Mean $\pm$ sd over three seeds; mix-NCE: InfoNCE with negatives from all recordings; best objective per backbone in bold.}
\label{tab:public}
\vspace{2pt}
\footnotesize
\setlength{\tabcolsep}{3.5pt}
\begin{tabular}{@{}lc@{\hskip 10pt}cccc@{\hskip 12pt}cccc@{}}
\toprule
 & & \multicolumn{4}{c}{PaPaGei-S} & \multicolumn{4}{c}{AnyPPG} \\
\cmidrule(lr){3-6}\cmidrule(l){7-10}
 & none & pooled & mix-NCE & \obj{fe-cos} & \obj{+fe-nce} & pooled & mix-NCE & \obj{fe-cos} & \obj{+fe-nce} \\
\midrule
\multicolumn{10}{@{}l}{\emph{held-out recordings: labels that vary within a recording}} \\
\quad SBP error (mmHg) & 10.97 & 8.93{\tiny$\pm$0.03} & 8.86{\tiny$\pm$0.02} & 8.77{\tiny$\pm$0.05} & \textbf{8.63}{\tiny$\pm$0.06} & 8.21{\tiny$\pm$0.01} & 8.16{\tiny$\pm$0.09} & 7.83{\tiny$\pm$0.04} & \textbf{7.79}{\tiny$\pm$0.03} \\
\quad DBP error (mmHg) & 5.48 & 4.81{\tiny$\pm$0.02} & 4.80{\tiny$\pm$0.00} & 4.73{\tiny$\pm$0.02} & \textbf{4.69}{\tiny$\pm$0.03} & 4.50{\tiny$\pm$0.02} & 4.45{\tiny$\pm$0.04} & 4.25{\tiny$\pm$0.02} & \textbf{4.22}{\tiny$\pm$0.02} \\
\quad HR error (bpm) & 5.90 & 2.50{\tiny$\pm$0.02} & 2.62{\tiny$\pm$0.02} & \textbf{2.29}{\tiny$\pm$0.01} & 2.50{\tiny$\pm$0.01} & 2.67{\tiny$\pm$0.02} & 2.84{\tiny$\pm$0.03} & \textbf{2.38}{\tiny$\pm$0.00} & 2.62{\tiny$\pm$0.02} \\
\quad state retrieval top-1 & 0.004 & 0.032{\tiny$\pm$0.001} & 0.123{\tiny$\pm$0.001} & 0.112{\tiny$\pm$0.001} & \textbf{0.196}{\tiny$\pm$0.003} & 0.040{\tiny$\pm$0.001} & 0.138{\tiny$\pm$0.001} & 0.137{\tiny$\pm$0.001} & \textbf{0.221}{\tiny$\pm$0.001} \\
\quad trait retrieval top-1 & 0.013 & 0.089{\tiny$\pm$0.013} & \textbf{0.218}{\tiny$\pm$0.025} & 0.116{\tiny$\pm$0.017} & 0.147{\tiny$\pm$0.022} & 0.089{\tiny$\pm$0.027} & \textbf{0.227}{\tiny$\pm$0.011} & 0.156{\tiny$\pm$0.017} & 0.164{\tiny$\pm$0.006} \\
\addlinespace[3pt]
\multicolumn{10}{@{}l}{\emph{held-out recordings: labels fixed per person}} \\
\quad age error (yr) & 12.15 & 10.23{\tiny$\pm$0.05} & 10.29{\tiny$\pm$0.02} & 10.22{\tiny$\pm$0.03} & \textbf{10.20}{\tiny$\pm$0.02} & 10.12{\tiny$\pm$0.11} & 9.93{\tiny$\pm$0.05} & 9.96{\tiny$\pm$0.08} & \textbf{9.91}{\tiny$\pm$0.07} \\
\quad sex AUROC & 0.500 & 0.558{\tiny$\pm$0.007} & \textbf{0.564}{\tiny$\pm$0.009} & 0.549{\tiny$\pm$0.007} & 0.546{\tiny$\pm$0.010} & 0.529{\tiny$\pm$0.017} & 0.547{\tiny$\pm$0.008} & 0.531{\tiny$\pm$0.004} & \textbf{0.552}{\tiny$\pm$0.005} \\
\addlinespace[3pt]
\multicolumn{10}{@{}l}{\emph{second database (MIMIC-III)}} \\
\quad state agreement & 0.000 & 0.049{\tiny$\pm$0.001} & 0.093{\tiny$\pm$0.001} & \textbf{0.129}{\tiny$\pm$0.003} & 0.094{\tiny$\pm$0.002} & 0.067{\tiny$\pm$0.004} & 0.101{\tiny$\pm$0.003} & \textbf{0.148}{\tiny$\pm$0.001} & 0.114{\tiny$\pm$0.002} \\
\quad HR error (bpm) & 4.60 & 2.49{\tiny$\pm$0.02} & 2.40{\tiny$\pm$0.06} & \textbf{2.02}{\tiny$\pm$0.01} & 2.24{\tiny$\pm$0.06} & 2.39{\tiny$\pm$0.05} & 2.47{\tiny$\pm$0.01} & \textbf{2.07}{\tiny$\pm$0.04} & 2.29{\tiny$\pm$0.06} \\
\quad SBP error (mmHg) & 8.26 & 8.62{\tiny$\pm$0.11} & 8.68{\tiny$\pm$0.04} & 8.47{\tiny$\pm$0.08} & \textbf{8.46}{\tiny$\pm$0.06} & 8.85{\tiny$\pm$0.16} & 8.69{\tiny$\pm$0.10} & 8.86{\tiny$\pm$0.06} & \textbf{8.67}{\tiny$\pm$0.03} \\
\bottomrule
\end{tabular}
\end{table*}

\subsection{Conditioning the loss raises state agreement}
\label{sec:ladder}

Fig.~\ref{fig:main}(a) places every run by the identity it memorises on
training recordings and the state it learns on held-out ones. The two trade off
across all 34 runs (Spearman $\rho=-0.79$), and the objectives order themselves
by how much of the state term is conditioned on the recording. State agreement
rises along that order on both backbones: from $0.136\pm0.006$ for the pooled regression, through
$0.195$ for cross-recording negatives and $0.240\pm0.002$ for mixed negatives
(mix-NCE), to
$0.307\pm0.001$ for \obj{fe-cos} (PaPaGei-S); the AnyPPG backbone gives the same
order and ends at $0.317\pm0.000$. Seed spread is at most $0.006$, against steps of
$0.05$--$0.08$ between rungs and $0.17$ in total.

The gain comes from the objective, not from the batch sampler. Drawing batches
of 96 different recordings with one window each, the composition a
cross-recording alignment loss would use, leaves the pooled objective at
$0.100$, and 24 recordings of 4 windows at $0.112$ (grey in
Fig.~\ref{fig:main}a); the $6\times16$ sampler
alone moves it to $0.136$. Of the $0.206$ between the cross-recording sampler
with a pooled loss and \obj{fe-cos}, the sampler accounts for $0.036$ and the
objective for $0.171$. Conditioning also cuts memorisation: trait agreement on
training windows drops from $0.49$--$0.69$ for the pooled loss to $0.23$--$0.32$
for \obj{fe-cos} (range over three seeds and both backbones), as
Eq.~\ref{eq:decomp} predicts once $\mu_r$ leaves the state gradient.

Nor is it a matter of regularisation strength. In the linear analogue of
Sec.~\ref{sec:fe} on ECSMP, with the ridge penalty held fixed along the whole $\lambda$
path (Fig.~\ref{fig:main}c), state agreement falls monotonically from the within
estimator to the pooled solution, from $0.130$ to $0.093$ on AnyPPG, a paired
difference of $+0.037$ $[0.033,0.040]$. Tuning each endpoint's penalty by
cross-validation shrinks the gap to $0.009$, so regularisation masks part of the
effect rather than creating it.

\subsection{The gain is state}
\label{sec:what}

\textbf{Labels that vary within a recording improve; labels fixed per person do
not.} Table~\ref{tab:public} reads the predicted embedding out with linear probes
against public labels, divided in advance into those that vary within a
recording (blood pressure, heart rate) and those fixed per person (age, sex). The
decomposition predicts a gain on the first group only, and that is what we
observe. \obj{fe-cos} lowers the error of every within-recording label on both
backbones (systolic pressure from $8.93$ to $8.77$\,mmHg on PaPaGei-S and from
$8.21$ to $7.83$ on AnyPPG, heart rate from $2.50$ to $2.29$\,bpm and from $2.67$
to $2.38$), while age and sex do not move. mix-NCE leaves blood pressure where the pooled loss left it and raises the
heart-rate error, so the gain comes from conditioning, not from contrast.

\textbf{Retrieval splits people from moments.} mix-NCE is best at trait retrieval, finding the right
person among 75 ($0.218$ top-1 against $0.116$ for \obj{fe-cos} on PaPaGei-S);
\obj{fe-cos}$+$\obj{fe-nce} is best at state retrieval, finding the right moment
among 256 windows of one person ($0.196$, against $0.123$ for mix-NCE and $0.032$
for the pooled loss).

\textbf{It is not shared artefact, and it is repeatable.} On 60 validation
recordings (one seed), injecting one band-limited $0.1$--$2$\,Hz motion-like waveform into both
the PPG and the ECG of a window leaves the gain of \obj{fe-cos} over the pooled
baseline at $+0.166$, against $+0.169$ clean and $+0.165$ with independent
waveforms; corrupting only the ECG leaves it at $+0.159$.
On 150 validation recordings, encoding each half of a window separately splits the target into a part that
repeats within the window and a part that does not: $+0.122$ $[0.116,0.129]$ of
the gain lies on the first against $+0.022$ $[0.019,0.025]$ on the second, and
\obj{fe-cos} reaches $94\%$ of the teacher's agreement with itself ($0.243$
against $0.260$, same covariates removed).

\subsection{Other sensors and another hospital}
\label{sec:external}

ECSMP~\cite{Gao_2021} is an independent database with chest ECG and wrist PPG from
different devices, in no pre-training corpus, aligned per recording with a
fitted clock model (52 recordings, 78\,h). Fig.~\ref{fig:main}(b) shows the same
ladder: $0.010\rightarrow0.072$ on PaPaGei-S and $0.028\rightarrow0.113$ on
AnyPPG. The gap between backbones is a domain-shift effect: PaPaGei-S reaches
a within-recording heart-rate tracking correlation of only $0.32$ on wrist PPG
against $0.78$ for AnyPPG, and the objective helps on both.

On 60 recordings of MIMIC-III, a second hospital that the bridge never saw,
state agreement rises from $0.049$ to $0.129$ on PaPaGei-S and from $0.067$ to
$0.148$ on AnyPPG, and the heart-rate error falls by $19\%$ and $13\%$
(Table~\ref{tab:public}). MIMIC-III PPG, like VitalDB PPG, is part of both
backbones' pre-training corpora~\cite{pillai2025papagei,Nie_2026}, so this test is external
to the bridge and to the objective, not to the backbones.

\subsection{Limitations}
\label{sec:limits}

White noise on the PPG lowers every objective, but the conditioned ones stay
ahead: at $0.2$ of the signal's sd, \obj{fe-cos} keeps $0.058$ and
\obj{fe-cos}$+$\obj{fe-nce} $0.093$ against $0.027$ for the pooled loss (same 60 recordings), so the conditional-logit form suits noisy input best. The gain lies
beyond heart rhythm and within what PPG can observe: mean RR is read equally well
by every objective ($\rho=0.69$--$0.70$), and the T/R amplitude ratio, to which
PPG has no path, by none ($\rho=0.04$--$0.05$). On MIMIC-III every
blood-pressure read-out stays above the zero-prediction error of $8.26$\,mmHg,
while state agreement and heart rate transfer.

\section{Conclusion}
\label{sec:conclusion}

A wearable is calibrated to its wearer and needs from an ECG teacher how that
wearer changes, yet pooled distillation spends much of its capacity on who they
are. Making the loss relative to each recording's own mean removes that component
exactly, a pooled anchor keeps it, and within-recording agreement with the
teacher more than doubles on two backbones and holds on two further databases. The raw alignment cosine cannot see
this, so ECG-guided PPG models should report within-recording agreement.

\vfill\pagebreak

\section{Compliance with Ethical Standards}
\label{sec:ethics}

This study was conducted retrospectively using de-identified human subject data
made available in open access by VitalDB and the MIMIC-III Waveform Database
Matched Subset, both through PulseDB~\cite{Wang_2023}, and by
ECSMP~\cite{Gao_2021}. Ethical approval was not required, as confirmed by the
licences attached to the open-access data.

\bibliographystyle{IEEEbib}
\bibliography{refs}

\end{document}